# New results on inconsistency indices and their relationship with the quality of priority vector estimation

Andrzej Z. Grzybowski

*Institute of Mathematics, Czestochowa University of Technology,*
*ul. Dabrowskiego 73,*
*42-201 Czestochowa, Poland*
andrzej.grzybowski@im.pcz.pl

**Abstract**
The article is devoted to the problem of inconsistency in the pairwise comparisons based prioritization methodology. The issue of "inconsistency" in this context has gained much attention in recent years. The literature provides us with a number of different "inconsistency" indices suggested for measuring the *inconsistency* of the pairwise comparison matrix (PCM). The latter is understood as a deviation of the PCM from the *consistent case* - a notion that is formally well-defined in this theory. However the usage of the indices is justified only by some heuristics. It is still unclear what they *really* "measure". What is even more important and still not known is the relationship between their values and the "consistency" of the decision maker's judgments on one hand, and the prioritization results upon the other. We provide examples showing that it is necessary to distinguish between these three tasks: the "measuring" of the "PCM inconsistency" and the PCM-based "measuring" of the consistency of decision maker's judgments and, finally, the "measuring" of the usefulness of the PCM as a source of information for estimation of the priority vector (PV). Next we focus on the third task, which seems to be the most important one in Multi-Criteria Decision Making. With the help of Monte Carlo experiments, we study the performance of various inconsistency indices as indicators of the final PV estimation quality. The presented results allow a deeper understanding of the information contained in these indices and help in choosing a proper one in a given situation. They also enable us to develop a new inconsistency characteristic and, based on it, to propose the PCM acceptance approach that is supported by the classical statistical methodology.

## 1. Introduction: an issue of inconsistency in pairwise comparisons.

One of the fundamental problems in decision making is the *prioritization* of available alternatives which is typically done by assigning a *priority weight* to each of them. The weights indicate the alternatives' relative importance with respect to a given criterion. The tuple of all priority weights forms a *priority vector* (PV) and deriving the PV on the basis of the information gathered from a decision maker (DM) the essence of all prioritization techniques. Many of these techniques are based on *pairwise comparisons* of the decision alternatives. As a result of such comparisons, a *pairwise comparison matrix* (PCM) is built - the elements of the PCM represent the DM judgments about the values of the priority weights' ratios. Although the idea of pairwise comparison is extremely natural and certainly very old, perhaps its first modern scientific applications were analyzed in (Fechner 1860).

Nowadays the pairwise comparison is a common technique that is primarily used in the analytic hierarchy process (AHP) - one of the most popular tools for multi-criteria decision making (MCDM). AHP was developed in the seventies and eighties of the last century by Thomas Saaty. Saaty's seminal study (Saaty1977) had an undeniably great impact on the development of the pairwise comparisons based prioritization methodology. Present-day applications of the AHP include such diverse problems as shipping management (e.g. Bulut et al. 2012), evaluation of new service concepts (e.g. Lee et al. 2012), or some military tasks (Jin & Rothrock 2010) to name just a few interesting examples from very recent years. Two theoretical issues connected with the usage of the pairwise comparisons are of special interest: the *choice of a prioritization technique* and *inconsistency evaluating*. The former refers to the PCM-based PV estimation methods, while the latter concerns "measuring" the credibility of the PCM (or of the DM her/himself) as a source of information about the PV. It is claimed (and it is quite intuitive) that serious errors in judgments about the priority ratios make the data contained in PCM useless and that they may result in poor PV estimates, see e.g. (Saaty 1980, 2004, Saaty &Vargas 1993). In decision making practice it is a very important problem. Therefore, in recent years, we are presented with a number of papers dealing solely with the analysis of the inconsistency of the PCM. According to this literature consistency control is nowadays "a unique and routine part of every AHP study" see (Bulut et al 2012), and the "possibility of evaluating, in a direct manner, the inconsistency of decision makers when eliciting the judgments" is of special importance in the AHP , see ( Aguarón et al. 2014, Altuzarra et al. 2010). The importance of the inconsistency measurement in the AHP practice was also emphasized in a number of application-oriented articles, (e.g. Duru et all 2012, Bulut et al 2012, Jin & Rothrock 2010, Lee et al. 2012) and/or in the context of group decision making, (e.g. Aguarón et al. 2014, Zhang et al. 2012, Lee 2012).

One can also find a number of articles devoted to the development of procedures enabling the consistency "improvement" and/or "monitoring", usually with the underlying aim of improving the final estimate quality, see e.g. (Benitez 2012, Bozóki et al. 2011, Koczkodaj &Szarek 2010, Lamata 2002, Liu et al. 2014, Xia et al. 2013, Saaty 2003, Saaty& Ozdemir 2003).

In order to "measure" the inconsistency of a given PCM, various characteristics (called indices) are proposed. As a matter of fact these indices are not any measures (certainly not in the mathematical sense). They are just some kind of characteristics of the degree of the PCM deviation from the one obtained in a perfect judgment case. The first and perhaps still the most popular inconsistency characteristic is due to Saaty. In the fundamental paper (Saaty 1977) he introduced an inconsistency index - denoted here as SI - which is closely related to his right eigenvalue prioritization method (REV). Another popular index is connected with a prioritization technique that is known as the Row Geometric Mean method (GM). The GM was introduced in a paper (Crowford & Williams 1985) In the same article the authors also suggested the *Geometric Consistency Index* (GI). The practical usage of this characteristic is analyzed e.g. in (Aguarón & Moreno-Jiménez 2003). Yet another interesting proposal is due to Koczkodaj. In (Koczkodaj1993) he proposed an inconsistency index (KI) that is based upon the notions of a *triad* and its inconsistency. Koczkodaj's index KI is not connected with any specific prioritization technique. Its performance was analyzed in various papers (e.g. Bozóki &Rapcsák 2008, Koczkodaj &Szwarc 2014).

Apart from the indices SI, GI and KI, we are also presented with various other PCM inconsistency characteristics ( e.g. Dijkasra 2013, Grzybowski 2010, 2012, Kazibudzki 2012,Pelaez and Lamata 2003) or (Dong et al. 2014) for interval PCMs. There are also some proposals for measuring consistency in the fuzzy pairwise comparison framework, such as the centric consistency index (which is based on GI) proposed by Bulut et al. (2012). However it seems undoubtful, that these three above-mentioned indices (SI, GI and KI) are the most widely used ones in the pairwise comparisons methodology, see e.g. ( Choo & Wedlay 2004, Lin 2007, Grzybowski 2012,Fedrizzi & Brunelli 2010,Dong et al. 2008).

All the inconsistency indices known from literature have one common feature: they are nonnegative and they equal 0 only in the case of a perfectly *consistent* PCM - a notion formally defined in this theory. The users of these indices also *hope* that greater index values indicate worse consistency of the DM judgments. In some problems it would be perhaps the most desired property of any inconsistency index. However such a claim is supported only by some heuristic arguments. One can find articles where such arguments are based on various intuitive "psychological" requirements, which according to the authors' opinions, should be reflected by the index properties. Among the literature we can even find some interesting attempts to construct a system of intuitive, psychologically-justified axioms which should be satisfied by "good" inconsistency indices, (Brunelli &Fedrizzi 2013, Koczkodaj & Szwarc 2014)

Another claim, fundamental for many applications, is the following: "the less consistent the DM judgments, the poorer are the PV estimates". It seems intuitive, but is it true? It turns out that it is not always - we provide examples here showing that the *improving* of the DM judgments *consistency* may lead to PV estimate's *errors increment*. Thus it is important to distinguish these two tasks :

- characterization of the dependence between the PCM and the consistence the DM judgments and
- characterization of the dependence between the PCM and the PV estimate's *errors*

To our best knowledge, all literature so far devoted to inconsistency analysis focuses on the first task or takes the existence of the desired dependencies (between index values, judgment consistency and magnitudes of PV estimation errors) for granted.

The first of the above tasks is certainly important in some situations, (e.g. Temesi 2011,Brunelli& Fedrizzi 2013). For example it is argued that " ... the more rational the judgments are, the more likely it is that the decision maker is a good expert with a deep insight into the problem ... ", (Brunelli& Fedrizzi 2013). However this task is related to the psychological analysis of the decision making process and it is beyond the scope of this paper (although we will address some such issues very briefly).

In this article we focus on the second task, which is of primary interest in MCDM. We will study the relation between the values of inconsistency indices and the quality of the PV estimates (reflected in the magnitude of estimates errors). In this context we feel that the name "inconsistency index" should be replaced with "estimates quality indicator". However, in our paper, we study inconsistency characteristics that are already well known from literature, so we preserve the traditional terminology. Nonetheless, it should be understood that we are primarily interested in studying how to characterize the usefulness of the PCM as a source of information for estimation of the PV. The results of such studies allow the DMs deeper understanding of the information contained in the indices and may help her/him to choose the one that is good (or the best) in a given situation.

Finally we also have to address another terminological issue here. In literature one can come across the terms: *consistency* index and *inconsistency* index. Both can be found even in the same text and both are used even for the same characteristic. For example Saaty &Vargas (1984) use term "inconsistency" index, while later (Saaty 2004, Saaty Ozdemir 2003) use the term "consistency index" (although he admits that de facto it indicates "inconsistency" of the PCM), Dijkstra (2014) uses the terms "consistency index" as well as "inconsistency measures", whilst e.g. in ( Brunelli & Fedrizzi 2013, Bozóki & Rapcsák 2008) or (Koczkodaj 1993 ) authors prefer the term "inconsistency index". Although consistency and inconsistency measuring are dual and interchangeable problems, we feel that the term "inconsistency index" better reflects what we want to study, so we use it in our text, as hereinbefore. This terminological issue may seem a bit too academic, but we think that it should be elucidated at this point.

The paper is organized as follows. In the next section some necessary definitions and facts related to our problem are provided. Section 3 is devoted to a more detailed insight into various types of "inconsistencies" and their relation with the quality of the PV estimates. In Section 4 we present the results of the simulation analysis of the relationship between the inconsistency indices and the magnitude of the PV estimation errors. The article is concluded with some remarks stated in Section 5.

## 2. Preliminaries

Let us consider $n$ decision alternatives that are to be ranked according to a given criterion. Deriving PV from a given PCM is to estimate priority weights $\mathbf{w}=(w_1,\ldots,w_n)$ on the base of the matrix $\mathbf{A}=[a_{ij}]_{n \times n}$, where the elements $a_{ij}$ of the matrix $\mathbf{A}$ are the DM judgments about the priority ratios $w_i/w_j$, $i,j =1,\ldots n$. Usually the priority weights $w_i$, $i=1,\ldots,n$, are nonnegative and normalized to unity: $\sum_i^n w_i = 1$. In the AHP practice the DM judgments are typically expressed in linguistic terms which are transformed to real positive values taken from an appropriate numeric scale.

**Definition 1**

A given matrix $\mathbf{A}=[a_{ij}]_{n \times n}$ is called *reciprocal* PCM (RPCM) if the condition $a_{ij}=1/a_{ji}$ holds for any $i,j =1,\ldots n$.

**Definition 2**

A given matrix $\mathbf{A}=[a_{ij}]_{n \times n}$ is called *consistent* PCM (or *cardinally* transitive) if it is reciprocal and its elements satisfy the condition: $a_{ij}a_{jk} = a_{ik}$ for all $i, j, k =1,\ldots,n$.

The necessary and sufficient condition for any positive matrix $\mathbf{A}$ to be consistent is the existence of a *certain* vector $\mathbf{w}$ satisfying $a_{ij}=w_i/w_j$ for all $i,j=1,\ldots,n$.

Over the last decades many prioritization methods have been proposed in literature. Choo et al. (Choo & Wedley2004) discussed 18 such methods. Lin (Lin 2007) argues, that among these methods we have only 15 actually different ones. In very recent years some new interesting proposals have appeared in the literature, see e.g. (Kazibudzki 2011,2012, Grzybowski 2010, 2012, Dijkistra 2013). All these prioritization methods are based on various concepts of the estimation criteria and/or on different assumptions about the sources of PCM inconsistency. However it is perhaps the REV which is the most popular. It was developed and applied in the AHP by Saaty, (Saaty 1977,1980). In his approach to obtain the estimate of the PV on the base of the matrix $\mathbf{A}$ we need to solve eigenvector equation

$$\mathbf{A}\mathbf{w}=\lambda_{\max}\mathbf{w} \qquad (1)$$

where $\lambda_{\max}$ is the principal eigenvalue. It is well-known (Perron-Frobenius theorem) that for any positive *reciprocal* matrix $\mathbf{A}$ the value $\lambda_{\max}$ is always real, unique and not smaller than $n$. In Saaty's approach as an estimate for the true PV, we assume the normalized right eigenvector $\mathbf{w}$ associated with this principal eigenvalue.

In reality, the PCM is usually inconsistent, even if the pairwise comparisons are done very carefully. So, as it was argued in the introduction, we need to measure the degree of the PCM deviation from the perfect case. According to Saaty's concept the inconsistency of the data contained in the PCM is measured by an inconsistency index SI($n$) that is computed according to the formula, (Saaty 1980):

$$SI(n) = \frac{\lambda_{\max} - n}{n - 1} \qquad (2)$$

In AHP practice, the value of SI is compared with an average consistency index ASI(*n*) which is computed from a sample of 500 randomly generated reciprocal matrices of order *n*. Saaty proposed the so-called consistency ratio CR=SI(*n*)/ASI(*n*) for testing whether the information contained in the PCM is consistent enough to be acceptable. Various shortcomings of this approach have been reported in literature, so a number of modifications have been proposed, (Alonso & Lamata 2006 , Grzybowski 2012)

Most of the REV alternatives are optimization based. Among them the most competitive and very popular is the GM, (see Budescu et al. 1986, Lee 2007, Choo & Wedley 2004, Grzybowski 2012, Kazibudzki 2012, Zahedi 1986). The estimates of the weights in the GM method are obtained from the following formula:

$$w_i = \left(\prod_{j=1}^{n} a_{ij}\right)^{1/n} \bigg/ \sum_{i=1}^{n} \left(\prod_{j=1}^{n} a_{ij}\right)^{1/n} \qquad (3)$$

There is also a consistency measure connected with the GM method. The index , denoted here as GI, was introduced by Crawford and Williams and is defined as follows, see (Crawford & Williams1985):

$$GI(n) = \frac{2}{(n-1)(n-2)} \sum_{i<j} \log^2(a_{ij} w_j / w_i) \qquad (4)$$

This index was put into the AHP practice mainly by Aguaron and Moreno-Jimenez (Aguaron & Moreno-Jimenez 2003). These authors also found formula describing the relation between *GCI* and *CI* and proposed consistency thresholds for acceptance of the PCM based on the value of *GCI*.

The third popular inconsistency index is due to Koczkodaj, (Koczkodaj 1993). To define the index we need the notion of a triad. For any three different decision alternatives we have three meaningful priority ratios - say $\alpha, \beta, \chi$ - which occupy different places in the a PCM **A**=[$a_{ij}$]$_{n \times n}$. The tuple ($\alpha,\beta,\chi$) is called a *triad* if $\alpha=a_{ik}$, $\beta=a_{ij}$, $\chi=a_{kj}$ for some different *i, j ,k* $\leq n$. It is obvious that for all triads in any *consistent* PCM the equality $\beta=\alpha\chi$ holds. Equivalently in such a case equations 1-$\beta/(\alpha\chi)$=0 and 1-$\alpha\chi/\beta$=0 have to be true. Therefore Koczkodaj proposed the following index *TI* for characterization of the triad's inconsistency:

$$TI(\alpha, \beta, \chi) = \min\left[\left|1 - \frac{\beta}{\alpha\chi}\right|, \left|1 - \frac{\alpha\chi}{\beta}\right|\right]$$

Now, Koczkodaj's inconsistency index *KI* of any reciprocal PCM is defined as a maximum of triad's inconsistencies i.e.:

$$KI = \max[TI(\alpha,\beta,\chi)] \qquad (5)$$

where the maximum is taken over all possible triads in the upper triangle of the PCM.

The three above-introduced inconsistency indices are used for characterization of the inconsistence of the PCM.

At this point we should realize that we have three actually different notions:

- the *PCM inconsistency* - understood as PCM deviation from the consistent case, see Definition 2, and expressed in terms of the specific inconsistency index values,

- the *DM inconsistency* - perhaps a better phrase should be DM *trustworthiness* - that is somehow reflected by both the number and the magnitude of his/her judgment errors.

- the *usefulness of the* PCM as a basis for PV estimation, "measurement" of which is the implicit yet perhaps most important task in MCDM.

In the next section we present examples that illustrate some important problems connected with the relationship between these three notions.

## 3. PCM inconsistency, DM inconsistency and the quality of PV estimates.

In this section we state some remarks concerning the relationship between the three notions indicated in the above section title. It is a proper moment to address the notion of the "quality of PV estimates". The notions of *good* and *poor* estimates are imprecise. Certainly they are related to the estimation' errors but even the notion of an "error" raises important question: what performance criterion (in statistics usually called *a loss function*) should be applied to inform the DM how big a mistake is related to specific estimates of the PV. Moreover the errors also depend on the adopted prioritization method (i.e. PV estimator). Therefore we need to clarify here what methods and criteria will be taken into account in our research. So, hereafter we consider the two previously-described prioritization methods: the REV and the GM. For measuring the errors we adopt the following loss functions:

average absolute error : $AE(\mathbf{v},\mathbf{w}) = \dfrac{1}{N}\sum_{i=1}^{N}|v_i - w_i|$  (6)

average relative error: $RE(\mathbf{v},\mathbf{w}) = \dfrac{1}{N}\sum_{i=1}^{N}\dfrac{|v_i - w_i|}{v_i}$  (7)

where $\mathbf{v}=(v_1,...,v_n)$ is the true PV while the $\mathbf{w}=(w_1,...,w_n)$ is its estimate received with the help of a given prioritization method.

The average absolute error AE is a rather common loss function that is used in a number of papers devoted to simulation analysis of the prioritization methods, (e.g. Zahedi 1986, Lee 2007, Dong et al. 2008, Grzybowski 2012, Kazibudzki 2011). However in our opinion the average relative error RE is also of special interest in many applications, especially where - as in AHP - we build some kind of hierarchy and then we develop the overall final ranking by the aggregation of the partial rankings. This type of errors seems to be also of special importance in the group decision making. It is because even "small" errors (in terms of absolute values) may significantly change the final rankings if they are big in relation to the true value. That is why we also consider this kind of loss function here.

Now let us start our considerations with the following example which is a kind of the mental experiment.

**Example 1**

Let us look at the problem where a DM needs to rank four decision alternatives. Let us assume that *we* know *his true* PV: $\mathbf{v}=(0.46, 0.25, 0.19, 0.10)$. The matrix of the true priority ratios (MPR) related to this vector $\mathbf{v}$ is the following:

$$MPR(v) = \begin{bmatrix} 1 & 1.84 & 2.421 & 4.6 \\ 0.543 & 1 & 1.316 & 2.5 \\ 0.413 & 0.76 & 1 & 1.9 \\ 0.217 & 0.4 & 0.526 & 1 \end{bmatrix}$$

Now let us assume that instead of this above perfect matrix, the DM produces the following PCM:

$$\mathbf{A} = \begin{bmatrix} 1 & \underline{3.14} & 2.421 & 4.60 \\ \underline{0.318} & 1 & 1.316 & \underline{2.625} \\ 0.413 & 0.76 & 1 & \underline{4.147} \\ 0.217 & \underline{0.381} & \underline{0.241} & 1 \end{bmatrix}$$

We see that he/she made *three* errors: the underlined entries $a_{12}$, $a_{24}$ and $a_{34}$ are erroneous. Table 1 presents the PV estimates obtained with the help of both prioritization methods on the basis of the matrix **A** along with related errors.

**Table 1**. The estimates of true PV based on the matrix **A** and their errors.

| Prioritization method | Estimate of PV | Errors | |
|---|---|---|---|
| | | AE | RE |
| REV | (0.495036, 0.208474, 0.219384, 0.0771063) | 0.0322 | 15,65% |
| GM | (0.496284, 0.209004, 0.217993, 0.0767189) | 0.0321 | 15.58% |

Now let us imagine that apart from the previous errors, the DM also made an *additional* one - the one indicated in parenthesis in the following matrix **B**:

$$\mathbf{B} = \begin{bmatrix} 1 & \underline{3.14} & 2.421 & 4.60 \\ \underline{0.318} & 1 & (1.944) & \underline{2.625} \\ 0.413 & (0.514) & 1 & \underline{4.147} \\ 0.217 & \underline{0.381} & \underline{0.241} & 1 \end{bmatrix}$$

The estimates and errors received in this case are presented in Table 2.

**Table 2**. The estimates of true PV based on the matrix **B** and their errors.

| Prioritization method | Estimate of PV | Errors | |
|---|---|---|---|
| | | AE | RE |
| REV | (0.490538, 0.233224, 0.199963, 0.0762749) | 0.0203 | 10.58% |
| GM | (0.495711, 0.23016, 0.197499, 0.0766303) | 0.0216 | 10.75% |

We can see - maybe unexpectedly for some readers - that the matrix **B** leads to better estimation results than matrix **A** which contains one error less. It is true for both prioritization methods and regardless the type of considered loss function. Apparently in this example, the less consistent DM judgments surpassingly lead to better estimates.

This example may look a bit strange for those who are used to PCMs with the entries filled in with numbers taken from a proper scale (this refers especially to AHP users). However, we can also check how this example works when we use such a scale. For this purpose let us adopt the Saaty's scale which contains the integers from 1 to 9 and their reciprocals. In such a case the *perfect* DM should round the elements of MPR($v$) to nearest value from the scale. As a result in our example we obtain the following rounded MPR (denoted as RMPR):

$$RMPR(v) = \begin{bmatrix} 1 & 2 & 2 & 5 \\ 1/2 & 1 & 1 & 3 \\ 1/2 & 1 & 1 & 2 \\ 1/5 & 1/3 & 1/2 & 1 \end{bmatrix}$$

Let us also round elements of the matrices **A** and **B**. We receive the following matrices **RA** and **RB**:

$$\mathbf{RA} = \begin{bmatrix} 1 & \underline{3} & 2 & 5 \\ \underline{1/3} & 1 & 1 & 3 \\ 1/2 & 1 & 1 & \underline{4} \\ 1/5 & 1/3 & \underline{1/4} & 1 \end{bmatrix}, \quad \mathbf{RB} = \begin{bmatrix} 1 & \underline{3} & 2 & 5 \\ \underline{1/3} & 1 & (2) & 3 \\ 1/2 & (1/2) & 1 & \underline{4} \\ 1/5 & 1/3 & \underline{1/4} & 1 \end{bmatrix}$$

Table 3 presents the results obtained in this situation. These results are even more amazing than in the previous (unrounded) case. In case of the GM method (which appears to be better in this example) both types of errors related to the "worse" matrix **RB** are about twice less than those related to the "better" matrix **RA**.

**Table 3**. The estimates of true PV based on the matrices **RA** and **RB** and their errors.

| Prioritization method | Estimate of PV | Errors | |
|---|---|---|---|
| | | AE | RE |
| | Results based on the matrix **RA** | | |
| REV | (0.480098, 0.204182, 0.242105, 0.0736147) | 0.0361 | 19.13% |
| GM | (0.47871, 0.204547, 0.243248, 0.0734945) | 0.0360 | 19.19% |
| | Results based on the matrix **RB** | | |
| REV | (0.476078, 0.247112, 0.204738, 0.0720718) | 0.0154 | 10.08% |
| GM | (0.47871, 0.243248, 0.204547, 0.0734945) | 0.0166 | 10.23% |

So, we see that results of the PV estimation which are based on the information gathered from *less trustworthy* DM (i.e. producing PCM with greater number of errors) *may* be better. Interesting question is: how about the behavior of inconsistence indices in this case? What should they indicate in our example; the fact that the matrix **RB** contains more errors (and the DM judgments are less consistent) or the fact that **RB** is a better basis for PV estimation?

Let us check how they actually behave here. Table 4 presents the values of the three considered inconsistency indices: *CI*, *GI*, and *KI*. For both matrices **RA** all the indices indicate the matrix **RB** as the less consistent one - and it is true as we know (in that sense, that it contains more judgment errors). However it may suggest that these indices are better as indicators of the consistence of the DM judgments than as indicators of the estimation quality. Later on we will address this issue more thoroughly.

**Table 4**. The values of the inconsistency indices for matrices **RA** and **RB**.

| Inconsistency index | PCM: | |
|---|---|---|
| | **RA** | **RB** |
| SI | 0.017 | 0.058 |
| GI | 0.068 | 0.228 |
| KI | 4/9 | 2/3 |

Now we address another interesting issue connected with the PCM inconsistence. A quite popular belief (at least with AHP practitioners) is that consistent PCM yields no estimation errors. However it is not true. The problem was addressed e.g. in (Temesi 2011) or (Choo & Wedley 2004) where authors distinguish between the "consistent" and "error-free" PCMs. Temesi in (Temesi 2011) states a remark (Proposition 5 therein) " It is possible to have a consistent PCM that is not error-free". The following example shows that it is even very likely to come across such a case.

**Example 2**

Let us consider a DM who should rank four decision alternatives and again lat us assume that *we know* his/her true PV: *v*=(0.35, 0.3, 0.2, 0.15). In this case the MPR(*v*) looks as follows:

$$MPR(v) = \begin{bmatrix} 1 & 7/6 & 7/4 & 7/3 \\ 6/7 & 1 & 3/2 & 2 \\ 4/7 & 2/3 & 1 & 4/3 \\ 3/7 & 1/2 & 3/4 & 1 \end{bmatrix}$$

Let us also assume that our DM is very trustworthy and, using the Saaty's scale, he/she produces the following PCM that is equal to RMPR(**v**):

$$PCM = RMPR(v) = \begin{bmatrix} 1 & 1 & 2 & 2 \\ 1 & 1 & 2 & 2 \\ 1/2 & 1/2 & 1 & 1 \\ 1/2 & 1/2 & 1 & 1 \end{bmatrix}$$

Let us note that this PCM is consistent! However apparently it is not error-free. But all inconsistency indices take on the value 0 and it suggests that we deal with an ideal situation, but it is not so - based on this PCM we make estimation errors. Indeed, in such a case every prioritization method gives the same PV estimate, here *w*=(1/3,1/3,1/6,1/6), and the errors are the following: AE(*v*,*w*)= 0.025 and RE(*v*,*w*)= 10.91%. It is quite an interesting observation that the rounding errors alone may lead to erroneous, yet consistent PCM.

As we emphasized in the introduction, our principal aim is to study the performance of the inconsistency indices as indicators of the PV estimation quality. In the light of the above examples, it may happen that a "poor", inconsistent matrix allows one to derive good estimates, and it is also possible that even a consistent matrix leads to quite significant estimation errors. But the *right question* that should be asked now is: *how likely is it*? Therefore, to achieve our goal of finding a good estimation quality indicator, we should ask questions in a statistical manner: given a fixed value of an inconsistency index *how often* can we obtain good estimates of PV and *how likely* in the same case do the estimates appear to be poor? And finally: which inconsistency index, if any, is the best one in answering these questions?

Consequently, in order to choose a proper inconsistency index (with our purpose in mind) we need to perform a statistical study. The necessity of the analysis of the error distribution is quite obvious in the AHP and indicated by some authors, (e.g. Dijkstra 2013). However, to our best knowledge, there are no such results reported in literature so far. The only way one can achieve it is via computer simulations. The results of such studies are presented in the next section.

## 4. Indices comparison – simulation frameworks and results

In the Monte Carlo experiments described in this section, we simulate the prioritization problems under various assumptions concerning the nature of the judgment errors. The most common and natural ones are the errors resulting from the limitations of the human brain capabilities. In literature devoted to simulation analysis of the prioritization methods, these errors have been modeled since the 1980's. They are usually treated as realization of random variables. In such a case the relation between the PCM elements and the true priority ratios is often expressed in the following form(e.g.Saaty 1980) :

$$a_{ij} = \varepsilon_{ij} \frac{w_i}{w_j} \tag{8}$$

Probability distributions (p.d.) of the perturbation factor $\varepsilon_{ij}$ mainly involve log-normal, gamma, uniform (e.g. Budescu et al 1986, Zahedi 1986, Basak 1998, Grzybowski 2010 ) and truncated normal ( e.g. Choo & Wedley 2004, Lin 2007, Grzybowski 2012). Apart from these most popular probability distributions, one can also find applications of the Laplace, Couchy, triangle and beta p.d. (for discussion see e.g. Dijcastra 2013 and Lipovetsky & Tichner 1996).

Additionally some authors (e.g. Bulut et al. 2012, Lipovetsky & Conclin 1996, Temesi 2011) argue that the errors can be also the result of :

- the questioning procedure itself,
- erroneous entering of the data (i.e. the judgment values),
- the scaling procedure (rounding errors)

The above types of mistakes can result in big errors in PCM - in (Lipovetsky & Conclin 1996), named as Unusual and False Observations - UFOs. Our Monte Carlo experiments take into account all these important situations.

Let us start our simulation analysis with a very simple experiment. In this experiment we investigate th*e impact of* the *magnitude of a single error* ε on the values of the indices under consideration.

For a given number of decision alternatives *n,* in this simulation framework we run the following steps:

*Step* 1. Randomly generate a priority vector *v* and related perfect MPR(*v*) = **M**

*Step* 2. Randomly choose a position (*i,j*) in **M,** $i \in \{1,...,n-1\}$ and $j \in \{i+1,...,n\}$

*Step* 3. Randomly choose a number ε from an interval [1.01, 1.075]

*Step* 4. Successively for each *k* from 1 to $N_e$ replace an element $m_{ij}$ with $m_{ij}\varepsilon^k$ and $m_{ji}$ with $1/(m_{ij}\varepsilon^k)$ . After *each* replacement, compute the values of examined indices as well as the REV and GM estimates of the vector *v* along with these estimates' errors AE and RE. As a result we obtain the vectors: VectorSI, VectorGI, VectorKI, VectorAE(REV), VectorAE(GM), VectorRE(REV), and VectorRE(GM). The dimension of each of these vectors equals $N_e$.

*Step* 5. Compute the Spearman and Pearson correlation coefficients between the vector of errors **e**={ε, $\varepsilon^2$,...,$\varepsilon^{Ne}$} and each of the vectors received in Step 4.

*Step* 6. Repeat Steps 1 to 5 $N_R$ times

*Step* 7. Return the *mean* values (w.r.t the number of runs) of all correlation coefficients computed during all runs in steps 5.

The above simulation framework will be addressed as *magnitude of a single error simulation framework* (MSE-SF).

We perform experiments based on MSE-SF for *n* = 4,..., 7. In each experiment the number of runs equals $N_R$ =1000 and the number of error increments equals $N_e$ =25 .

In the analysis of inconsistency indices the most important is the Spearman rank correlation coefficient ρ. This coefficient tells us whether the increment of a single judgment error results in an increment of a given index value. It is one of the desired features of a good inconsistency indicator. Consequently, any inconsistency index should have the value of ρ equal to 1 (at least very close to). It appears that all considered indices perfectly pass this trial - all of them have *in each* run the value of the Spearman correlation coefficient equal to 1 and, obviously, such were the means of them. Moreover, it turns out that under the MSE-SF the Spearman correlation coefficient between the magnitude of *estimation errors* and the *indices values* in each case also equals 1. This was observed regardless of the estimation method (REV or GM) and the type of the loss function (AE or RE).

The Pearson correlation coefficients $r$ are less important in the analysis of inconsistency indices. However they are quite interesting, because they tell us whether the changes of the values of one quantity results in proportional changes in values of another (or: whether the relation between these quantities is close to linear one). In the case of the estimates errors and the inconsistency indices such relationship would be very convenient, especially from the point of view of constructing PCM acceptance rules. It appears that under the MSE-SF all considered indices are very satisfactory, also with respect to such a requirement. Although in each case the highest values of Pearson correlation $r$ with estimation errors are gained by the index KI, all of the indices perform really well having value $r$ greater than 0.965 ( again regardless of the estimation method and the type of the loss function). On the other hand, the Pearson correlation coefficients between the inconsistency indices and the magnitude of the judgment errors are higher in the case of the indices GI and SI, although again *all* the coefficients are very high ( greater than 0.960). Nonetheless it *may* suggest that the index which is better in indicating DM inconsistency, can be worse in indicating the PCM's fitness for estimation of the PV. But as a whole, the results received under the MSE-SF do not reveal any important differences between the examined indices.

Let us check whether this observation will be confirmed under other simulation frameworks.

The second natural task is to investigate the relationship between the inconsistency indices and the *number of equal errors*. So now we describe an experiment designed to study this relationship.

But first let us introduce a ***new inconsistency characteristic***. The new index is based upon Koczkodaj's idea. Koczkodaj's index *KI*, is defined as the *maximum* of all triad inconsistencies *TI(a,b,c)* , see (5). It is obvious that the value of *KI* is rather robust against the changes in the number of errors if their magnitude is the same (the influence of changing number of errors is shadowed by the biggest one). Thus we propose another characteristic which is much more sensitive to such changes in PCM: the average value of all "triad inconsistencies". More precisely, the new index *ATI* is defined by the formula:

$$ATI = \text{Mean}[TI(\alpha,\beta,\chi)]$$

where the above arithmetic mean is computed on the basis of all different triads $(\alpha,\beta,\chi)$ in the upper triangle of the considered PCM.

Obviously if there is only one judgment error in the PCM then the new index ATI takes exactly the same value as KI, and thus under the MSE-SF both of them perform exactly the same.

The simulation framework for the study of the relationship between the consistency indices and the *number of equal errors* is the following ( this framework will be denoted NEE-SF).

For a given number of decision alternatives $n$ in this simulation framework we run the following steps:

*Step* 1. Randomly generate a priority vector $v$ and related perfect MPR($v$) = **M**

*Step* 2. Choose random permutation *p* of all elements $m_{ij}$ in the upper triangle of **M**

*Step* 3. Choose random value $\varepsilon_r$ of the error $\varepsilon$ from the interval [1.1,..., 1.8]

*Step* 4. Successively (in the order defined by the permutation *p*) replace each element $m_{ij}$ in the upper triangle with $m_{ij}\varepsilon_r$ and then $m_{ji}$ with $1/(m_{ij}\varepsilon_r)$. Similarly as in the framework MSE-SF, after each such pair of replacements compute the values of all examined indices as well as the REV and GM estimates of the vector *v* along with these estimates' errors AE and RE. We obtain the vectors: VectorSI, VectorGI, VectorKI, VectorATI, VectorAE(REV), VectorAE(GM), VectorRE(REV), and VectorRE(GM). The dimensions of these vectors equal $n(n-1)/2$.

*Step* 5. Compute the Spearman and Pearson correlation coefficients between the number of errors - i.e. the vector $(1, 2, ... , n(n-1)/2)$ - and each of the vectors returned in Step 4.

*Step* 6. Repeat Steps 2 to 5 $N_p$ times

*Step* 7. Repeat Steps 1 to 6 $N_R$ times

*Step* 8. Return the arithmetic *mean* values (w.r.t the number of all runs) of all correlation coefficients computed during all runs in steps 5.

Note, that Step 6 is important, because for a given PCM the correlation between the vectors computed in Step 4 and the number of errors may depend on both the order of disturbed elements (i.e. on the permutation *p*) as well as on the magnitude of the error randomly drawn in step 3. Thus in our experiments for each perfect matrix **M** generated in Step 1 we observe the examined relationship in $N_r$ different setups.

We run simulation experiments based on NEE-SF for $n = 4,..., 7$. The studies for $n=3$ are not interesting, because for such a number of alternatives the considered indices are directly related to each other, (Bozóki & Rapcsák 2008, Dijkstra 2013). For each considered number of alternatives *n* the number of $N_p$ equals 5 and $N_R$ equals 200. Consequently all mean values of considered correlation coefficients are based on 1000 different random setups.

As opposed to the previous results, the ones obtained under the NEE-SF are not so predictable. They reveal some interesting findings. First, let us look at the Spearman rank correlation between the number of judgment errors and the magnitude of the PV estimation errors. Their values are presented in Table 5. All these rank coefficients are significantly less than 1. It shows that such phenomena that was described in our Example 1 is not very rare - it is not so unusual when an additional judgment error inflicted on the PCM results in diminution of the PV estimation error

**Table 5**. Average Spearman rank correlation coefficients between the number of equal multiplicative judgment errors and the corresponding estimates' errors AE(REV), RE(REV), AE(GM), RE(GM). Results based on 1000 random setups for each $n=4,5,6,7$.

|         | $n=4$ | $n=5$ | $n=6$ | $n=7$ |
|---------|-------|-------|-------|-------|
| AE(REV) | 0.812 | 0.843 | 0.863 | 0.877 |
| RE(REV) | 0.869 | 0.884 | 0.897 | 0.906 |
| AE(GM)  | 0.847 | 0.879 | 0.894 | 0.909 |
| RE(GM)  | 0.868 | 0.898 | 0.914 | 0.927 |

Another important observation is connected with the Spearman rank correlation coefficients between the inconsistency indices and the PV estimation errors. Again, none of them equals 1, not even once. Their average values are presented in Table 6. It can be observed that they are even not close to 1. This fact proves that it is quite possible to face the situation where the inconsistency index values are misleading - they may indicate a given PCM as a good one, while it turns out to be poor as the PV estimation basis (this remark refers to all investigated indices). However it can also be seen that in all cases (i.e. regardless of the number of alternatives, the prioritization method and the type of the loss function) the new index ATI has the highest rank correlation with all considered characteristics of the PV estimation quality as well as with the number of inflicted judgment errors. The presented results also confirm that the index KI can be very misleading in the considered framework i.e. if the judgment errors have equal or very similar magnitudes (we have expected such performance of KI, and it was the reason for introducing the new index ATI).

Table 7 contains values of the Pearson correlation coefficient. As we have already mentioned, it is not the most important, yet an interesting indicator of the inconsistency indices performance. We see that also these results confirm the dominance of the ATI under the NEE-SF.

**Table 6**. Average Spearman rank correlation coefficients between the values of an indicated index (in a column head) and the number of equal multiplicative judgment errors (NE) and the corresponding estimates' errors AE(REV), RE(REV), AE(GM), RE(GM). Results based on 1000 random setups for each $n=4,5,6,7$.

|         | SI    | GI    | KI     | ATI   | SI    | GI    | KI     | ATI   |
|---------|-------|-------|--------|-------|-------|-------|--------|-------|
|         |       | $n=4$ |        |       |       | $n=5$ |        |       |
| NE      | 0.512 | 0.495 | -0.025 | 0.627 | 0.623 | 0.607 | -0.008 | 0.702 |
| AE(REV) | 0.232 | 0.214 | -0.175 | 0.362 | 0.371 | 0.351 | -0.126 | 0.460 |
| RE(REV) | 0.297 | 0.280 | -0.139 | 0.423 | 0.424 | 0.407 | -0.098 | 0.509 |
| AE(GM)  | 0.285 | 0.249 | -0.135 | 0.396 | 0.438 | 0.421 | -0.088 | 0.525 |
| RE(GM)  | 0.298 | 0.259 | -0.131 | 0.404 | 0.452 | 0.433 | -0.087 | 0.536 |
|         |       | $n=6$ |        |       |       | $n=7$ |        |       |
| NE      | 0.687 | 0.676 | 0.001  | 0.758 | 0.735 | 0.727 | 0.005  | 0.798 |
| AE(REV) | 0.466 | 0.452 | -0.090 | 0.547 | 0.538 | 0.529 | -0.068 | 0.611 |
| RE(REV) | 0.514 | 0.501 | -0.067 | 0.591 | 0.579 | 0.571 | -0.050 | 0.648 |
| AE(GM)  | 0.528 | 0.516 | -0.059 | 0.606 | 0.599 | 0.591 | -0.042 | 0.668 |
| RE(GM)  | 0.545 | 0.532 | -0.056 | 0.622 | 0.614 | 0.605 | -0.040 | 0.682 |

**Table 7**. Average Pearson correlation coefficients between the values of an indicated index (in a column head) and the number of equal multiplicative judgment errors (NE) and the corresponding estimates' errors AE(REV), RE(REV), AE(GM), RE(GM). Results based on 1000 random setups for each $n=4,5,6,7$.

|         | SI    | GI    | KI    | ATI   | SI    | GI    | KI    | ATI   |
|---------|-------|-------|-------|-------|-------|-------|-------|-------|
|         |       | $n=4$ |       |       |       | $n=5$ |       |       |
| NE      | 0.582 | 0.586 | 0.172 | 0.663 | 0.675 | 0.680 | 0.184 | 0.745 |
| AE(REV) | 0.359 | 0.368 | 0.024 | 0.456 | 0.470 | 0.475 | 0.060 | 0.555 |
| RE(REV) | 0.374 | 0.378 | 0.033 | 0.470 | 0.489 | 0.494 | 0.070 | 0.574 |
| AE(GM)  | 0.360 | 0.364 | 0.030 | 0.457 | 0.492 | 0.496 | 0.081 | 0.572 |
| RE(GM)  | 0.385 | 0.390 | 0.047 | 0.482 | 0.512 | 0.517 | 0.092 | 0.594 |
|         |       | $n=6$ |       |       |       | $n=7$ |       |       |
| NE      | 0.733 | 0.737 | 0.200 | 0.794 | 0.774 | 0.777 | 0.211 | 0.826 |
| AE(REV) | 0.548 | 0.552 | 0.095 | 0.620 | 0.606 | 0.610 | 0.127 | 0.669 |
| RE(REV) | 0.568 | 0.572 | 0.105 | 0.641 | 0.625 | 0.628 | 0.137 | 0.688 |
| AE(GM)  | 0.575 | 0.578 | 0.118 | 0.643 | 0.637 | 0.640 | 0.151 | 0.695 |
| RE(GM)  | 0.595 | 0.599 | 0.128 | 0.664 | 0.654 | 0.658 | 0.158 | 0.714 |

Tables 6 and 7 reveal also other interesting findings about the indices. First, the greater the number $n$ of considered decision alternatives is, the greater their correlation with the PV estimation quality- and it seems that it grows monotonically. Second, all considered indices demonstrate the higher correlation with the relative errors than with the absolute ones, and all indices have the highest correlation with the errors RE(GM). The differences in correlations are not too impressive but they are statistically significant and - what we emphasize here -they occur in all cases of examined numbers of alternatives (although we present results here for n=4,..,7 we have also carried such studies for n=8,9,10.)

The two above simulation experiments were designed for studying two relationships: between index values and *a*) the magnitude of exactly one judgment error as well as *b*) the number of errors (of the same magnitude). In these two special cases we have also studied the dependence of the estimation errors and the indices values. However in a real world situation one can rarely expect these two situations to occur. We rather expect that typically there are more errors than just one, and it is very likely that these errors have a different magnitude. It is obvious that the quality of the final estimates depends on both the number the errors and the magnitude of the errors.

To study this issue more deeply let us considered another simulation framework designed for modeling such, more realistic, situations.

Following the idea presented in (e.g. Choo & Wedley 2004, Lin 2007 and Grzybowski 2012) we consider the framework where the generated PCMs contain many small errors of different magnitude as well as - possibly - one large error placed at a random position in the PCM. All generated and disturbed PCMs are finally rounded to a given scale (a typical procedure in the AHP). Such simulation experiments consist of the following steps.

*Step* 1. Randomly generate a priority vector $v$ and related perfect MPR($v$) = **M**
*Step* 2. Randomly choose an element $m_{i0j0}$ in the upper triangle and replace it with $m_{i0j0} \varepsilon_B$ where $\varepsilon_B$ is a "big" error which is randomly drawn from the interval $D_B$.
*Step* 3. For each other element $m_{ij}$, $i<j\leq n$, randomly choose value $\varepsilon_{ij}$ of the small error according the p.d. $\pi$. Then replace the element $m_{ij}$ with $m_{ij}\varepsilon_{ij}$.
*Step* 4. Round all values in the upper triangle to the closest value from a considered scale.
*Step* 5. Replace all elements in the lower triangle of the PCM with the reciprocities of the appropriate elements from the upper triangle.
*Step* 6. After all replacements are done compute the values of all examined indices as well as the estimates of the vector $v$ along with these estimates' errors AE and RE. Remember values computed in this step as one record.
*Step* 7. Repeat Steps 2 to 6 $N_M$ times
*Step* 8. Repeat Steps 2 to 7 $N_R$ times
*Step* 9. Return *all* records organized as one database.

The above simulation framework is denoted as MSOBE - SF.

In our experiments the p.d. $\pi$ of the small error in Step 3 is one of the following four most frequently considered in literature types of p.d. : gamma, log-normal, truncated normal, and one that is uniform. In each case the parameters of the distributions are set in such a way that their expected values equal 1. The support of the truncated normal and the uniform distribution is the interval $D_S$=[0.5, 1.5]. In case of the remaining two distributions, their parameters were prescribed in a way ensuring that the probability of the interval $D_S$ is greater than 0.98.

Let us note that in the above simulation framework by generating randomly disturbed PCMs (Steps 1-6) we also "generate" four types of *random* errors related to these matrices - namely AE(REV), AE(GM), RE(REV) and RE(GM). Now we are going to study the distributions of these random errors and their relationship with the values of the inconsistency indices. These distributions are of our primary interest because, as we have already argued, it may always happen that a "poor" matrix results in good "estimates", and it is also possible that we face opposite phenomena. But sometimes the chances for a big estimate error are "small", in other situations the chances are "significantly big", and we hope to find a way to distinguish between such cases with the help of the inconsistency indices. To study the usefulness of the indices in such a context we analyze the relationship between their values and the *quantiles* of the estimates' errors distributions. For this purpose in our analysis the whole simulation database (returned in Step 9 of MSOBE-SF) is sorted according the values of a given index and then split into $N_C$ separate classes $IC_i$, ($i=1,...,N_C$). For each $i=1,...,N_C$, the class $IC_i$ is strictly connected with a unique subset of the generated PCMs - namely these which have the index values belonging to $IC_i$. On the other hand, each such subset of PCMs "produces" sets of random estimation errors (of all considered types). We will say: random errors related to the class $IC_i$, $i=1,...,N_C$. Let $SA_i(PM)$ and $SR_i(PM)$ denote the set of absolute and relative errors, respectively, related to the class $IC_i$, and received when using the prioritization method PM (in our research the REV or GM).

Here and later in our analysis, to ensure some kind of objectivity, for all considered inconsistency indices we use the same computer procedure for splitting the whole range of their observed values into *n* separate classes. The procedure is the following: the first class is from 0 to the quantile of order $1/n$, the last *n*-th class starts from the quantile of order $1-1/n$ to infinity. All remaining *n*-2 classes have the same length and cover the whole interval between these two quantiles.

For each set of error values $SA_i(PM)$ and $SR_i(PM)$, $i=1,...,N_C$, we compute the quantiles of the order 0.1, 0.5 (median) 0.9 as well as their arithmetic mean.

In case of a good *inconsistency* index one can expect that quintiles of any order should monotonically *increase* along with the ranges of the index values $IC_i$, $i=1,...,N_C$ (or - more technically - when the subscript *i*, increases from 1 to $N_C$ ). The same relation should be observed in the case of the mean of the error values. Any significant violation of such a relationship would contradict the considered idea of the inconsistency index and its usefulness would be questionable.

As an example illustrating this point, let us consider Table 8. This table presents results obtained for Saaty's index SI under the framework MSOBE-SF in case where *n=4*, and the perturbation factors $\varepsilon_{ij}$ (in Step 3) is generated -in equal proportions- according to all four considered probability distributions. The "big error" - generated in Step 2 - has uniform distribution on the interval $D_B=[2,4]$. Such errors are usually considered as "big", (e.g. Dijkastra 2013, Grzybowski 2012, Lee 2007). However in 1/4 of the simulation runs Step 2 is omitted, so 25% of records in the simulation database are related to cases with no "big errors". The presented results are based on 240 000 random reciprocal PCMs. The second column of Table 8 contains separate classes $IC_i$, $i=1,...,15$, of the SI values. The third one shows the arithmetic mean of the index in a given class. The next columns contain values of indicated quintiles in the appropriate sets of observed errors $SA_i(REV)$ as well as the arithmetic mean of these errors (the last column).

**Table 8 Performance of the index SI.** Statistical characteristics of the distribution of random errors AE(REV) related to different classes $IC_i$, $i=1,...,15$ of SI values. Results based on 240 000 random reciprocal PCMs obtained under the MSOBE-SF for $n=4$. The perturbation factors $\varepsilon_{ij}$ was generated - in equal proportions- according to gamma, log-normal, truncated normal, and uniform distributions.

| $i$ | classes $IC_i$ of SI | Mean SI in $IC_i$ | $p$-quantiles in $SA_i$(REV): $p=0.1$ | $p=0.5$ | $p=0.9$ | Mean AE in $SA_i$(REV) |
|---|---|---|---|---|---|---|
| 1  | 0.000 ÷ 0.0052  | 0.0025 | 0.0058 | 0.0143 | 0.0349 | 0.0181 |
| 2  | 0.0052 ÷ 0.016  | 0.0107 | 0.0083 | 0.0206 | 0.0473 | 0.0246 |
| 3  | 0.016 ÷ 0.027   | 0.0212 | 0.0115 | 0.0271 | 0.0549 | 0.0312 |
| 4  | 0.027 ÷ 0.039   | 0.0330 | 0.0136 | 0.0309 | 0.0621 | 0.0350 |
| 5  | 0.039 ÷ 0.050   | 0.0443 | 0.0152 | 0.0330 | 0.0692 | 0.0381 |
| 6  | 0.050 ÷ 0.061   | 0.0555 | 0.0161 | 0.0346 | 0.0722 | 0.0401 |
| 7  | 0.061 ÷ 0.072   | 0.0664 | 0.0166 | 0.0362 | 0.0744 | 0.0416 |
| 8  | 0.072 ÷ 0.083   | 0.0774 | 0.0169 | 0.0366 | 0.0760 | 0.0420 |
| 9  | 0.083 ÷ 0.094   | 0.0883 | 0.0165 | 0.0363 | 0.0746 | 0.0416 |
| 10 | 0.094 ÷ 0.105   | 0.0993 | 0.0166 | 0.0366 | 0.0748 | 0.0417 |
| 11 | 0.105 ÷ 0.117   | 0.1107 | 0.0166 | 0.0364 | 0.0724 | 0.0410 |
| 12 | 0.117 ÷ 0.128   | 0.1222 | 0.0165 | 0.0360 | 0.0710 | 00405 |
| 13 | 0.128 ÷ 0.139   | 0.1333 | 0.0163 | 0.0363 | 0.0704 | 0.0405 |
| 14 | 0.139 ÷ 0.150   | 0.1442 | 0.0162 | 0.0365 | 0.0671 | 0.0397 |
| 15 | 0.150 ÷ ∞       | 0.3012 | 0.0171 | 0.0360 | 0.0626 | 0.0388 |

From this table, for example, we see that in the subset of randomly disturbed PCMs for which the index SI has values in the interval $IC_8$=[0.072, 0.083) the quantile of order 0.1 in the set $SA_8$(REV) equals 0.0169. It tells us that 10% of all PCMs in this group allow us to estimate the true vector **v** with an absolute error AE not greater than 0.0169, while at the same time 90% of the PCMs results in errors not less than this value (remember that the errors analyzed in Table 8 are related to the REV prioritization method). The interpretation of the quantile of order 0.90 is analogous: 90% of PCMs in this subset yield errors less than 0.0760 while 10% of them yield errors which are not less than that. We may also put it in the following way: if we have PCM with the value of SI in the interval $IC_8$ then there is a 10% chance that the error AE is less than 0.0169 *and at the same time*, there is a 10% chance that this error is bigger than 0.076. We can also see that, roughly speaking, the mean error in such situation is 0.042 (the last column). *If the index SI was a good indicator* of the estimates' quality then for all subsets of PCMs related to classes $IC_i$, $i > 8$, the quantiles as well as the mean errors *should be greater*. But it is not true. For example for all classes $IC_i$, $i=9,...,14$ the quantiles of order 0.1 are less than 0.0169 (the quantile in the class $IC_8$), suggesting that the chances for small errors are greater when the index value increases. As a matter of fact, from Table 8 we see that the index *SI* tells us very little about the possible magnitude of the absolute estimation error AE(REV). For all classes $IC_i$, $i >6$ the empirical mean of the error AE changes very little, and what is worse, non monotonically - so the information contained in SI may be misleading! The same remark applies to its relation with all considered quantiles. Perhaps it can be even better seen from Fig. 1 which is an illustration of the Table 8. This figure consists of 4 scatter plots presenting the relationship between a given statistical characteristic of the errors AE(REV) in particular classes and the mean values of the index SI computed for each class $IC_i$, $i=1,...,15$ (these means are presented in the third column of Table 8).

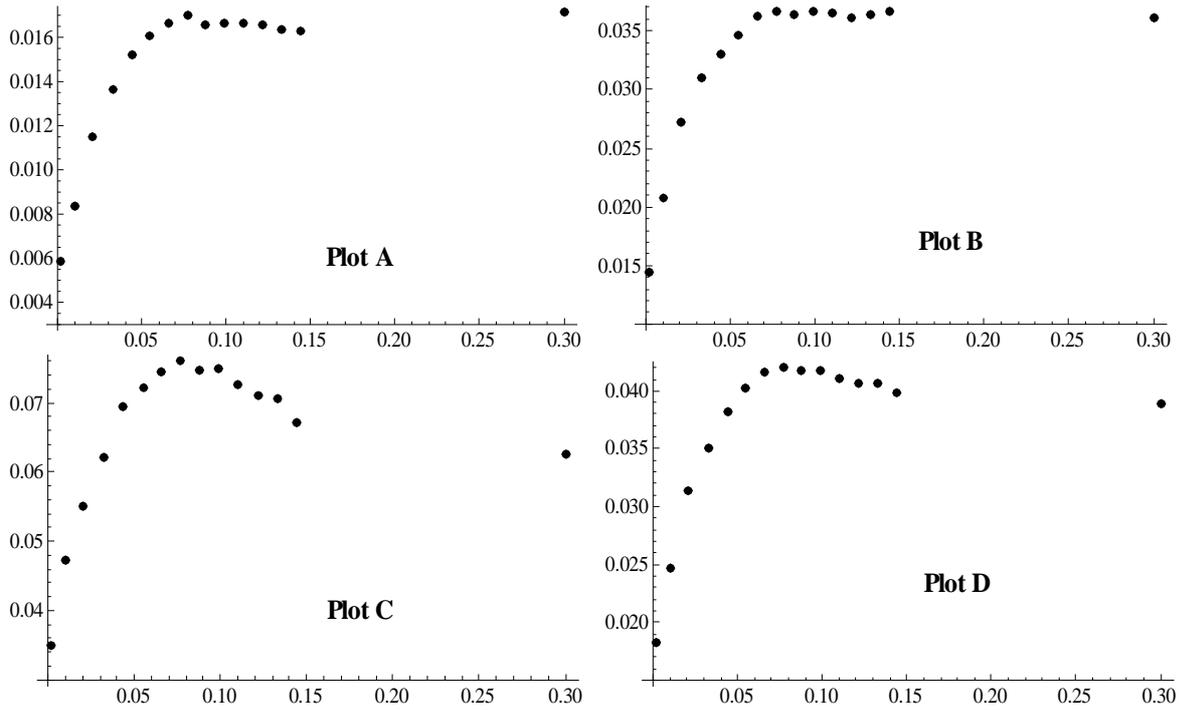

**Fig. 1. Performance of the index SI.** The relation between mean values of SI in $IC_i$ and the statistical characteristics of sets $SA_i(REV)$, $i=1,...,15$. Scatter plot of mean values of SI in successive classes of its values vs. quantiles of order 0.1 - Plot A, medians - Plot B, quantiles of order 0.9 - Plot C, and the means of the error AE(REV) - Plot D. Plots are based on 240 000 random reciprocal PCMs generated for $n=4$.

Now, it is interesting whether the other indices behave the same. Let us analyze Table 9 and Fig. 2. They present the results describing the performance of the index ATI whose values have been computed in exactly the same experiment. Now we see that the relationship between the mean value of ATI (in the classes $IC_i$) and the related quantiles and mean values in $SA_i(REV)$ is almost perfectly monotonic. It can be clearly seen in Fig. 2 which, similar as Fig. 1, consists of 4 scatter plots illustrating the relationship between a given statistical characteristic of the sets $SA_i(REV)$ and the mean values of the index ATI computed for each class $IC_i$, $i=1,...,15$. The only characteristic which is not perfectly monotonic with respect to the index values is the quintile of order 0.9. However, even in this case, the monotonicity is not strongly violated and only in the intervals containing the highest values of the index ATI.

Another interesting and important difference between these two indices is that the sets of the SI values contain many outliers - in each of the plots placed in Fig 1 we can see how far from the other points lies the last one. It is because above 5% of all generated PCMs have very "big" and misleading values of the index SI, while the same PCMs have "normal" values of the index ATI. And for these, sometimes extremely big values of SI, the related PCM often appears to be not such a bad source of the information about the true PV.

**Table 9 Performance of the index ATI.** Statistical characteristics of the distribution of random errors AE(REV) related to different classes $IC_i$, $i=1,...,15$ of ATI values. Results based on 240 000 random reciprocal PCMs obtained under the MSOBE-SF for $n=4$. The perturbation factors $\varepsilon_{ij}$ was generated - in equal proportions- according to gamma, log-normal, truncated normal, and uniform distributions.

| $i$ | classes $IC_i$ of ATI | Mean ATI in $IC_i$ | $p$-quantiles in $SA_i(REV)$: $p=0.1$ | $p=0.5$ | $p=0.9$ | Mean AE in $SA_i(REV)$ |
|---|---|---|---|---|---|---|
| 1 | 0.000 ÷ 0.173 | 0.1111 | 0.0062 | 0.0152 | 0.0377 | 0.0191 |
| 2 | 0.173 ÷ 0.207 | 0.1867 | 0.0073 | 0.0182 | 0.0432 | 0.0223 |
| 3 | 0.207 ÷ 0.240 | 0.2230 | 0.0086 | 0.0217 | 0.0483 | 0.0256 |
| 4 | 0.240 ÷ 0.274 | 0.2614 | 0.0096 | 0.0235 | 0.0517 | 0.0280 |
| 5 | 0.274 ÷ 0.308 | 0.2895 | 0.0100 | 0.0251 | 0.0545 | 0.0294 |
| 6 | 0.308 ÷ 0.341 | 0.3251 | 0.0127 | 0.0302 | 0.0612 | 0.0344 |
| 7 | 0.341 ÷ 0.375 | 0.3586 | 0.0138 | 0.0312 | 0.0643 | 0.0358 |
| 8 | 0.375÷0.409 | 0.3918 | 0.0145 | 0.0322 | 0.0680 | 0.0373 |
| 9 | 0.409÷0.443 | 0.4260 | 0.0150 | 0.0336 | 0.0709 | 0.0388 |
| 10 | 0.443÷0.476 | 0.4593 | 0.0154 | 0.0346 | 0.0728 | 0.0398 |
| 11 | 0.476÷0.510 | 0.4925 | 0.0161 | 0.0348 | 0.0717 | 0.0400 |
| 12 | 0.510÷0.544 | 0.5264 | 0.0163 | 0.0354 | 0.0726 | 00405 |
| 13 | 0.544÷ 0.577 | 0.5597 | 0.0165 | 0.0363 | 0.0727 | 0.0412 |
| 14 | 0.577÷ 0.611 | 0.5930 | 0.0173 | 0.0369 | 0.0710 | 0.0413 |
| 15 | 0.611 ÷ ∞ | 0.6738 | 0.0194 | 0.0380 | 0.0682 | 0.0420 |

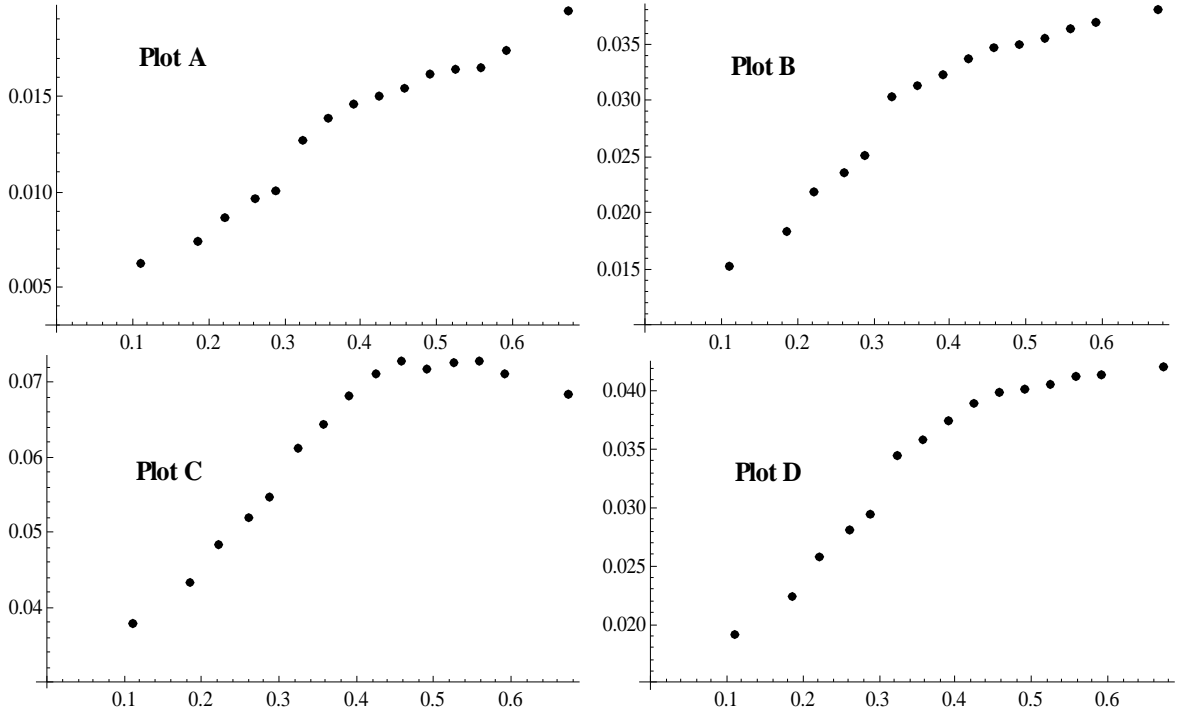

**Fig. 2. Performance of the index ATI.** The relation between mean values of ATI in $IC_i$, and the statistical characteristics of sets $SA_i(REV)$, $i=1,...,15$. Scatter plot of mean values of ATI in successive classes of its values vs. quantiles of order 0.1 - Plot A, medians - Plot B, quantiles of order 0.9 - Plot C, and the means of the error AE(REV) - Plot D. Plots are based on 240 000 random reciprocal PCMs for $n=4$.

The results presented so far concern only the loss function AE and the prioritization method REV. Now we provide results concerning other types of estimation errors. However we do not present any tables or figures like these two before - it would take up too much article space. Instead we present the Spearman rank correlation coefficients between the quantities presented in Table 8 (and illustrated in Fig. 1). As we indicated before, it is essential that in the case of a good inconsistency index its Spearman rank coefficient with all quantiles and means should be close to 1.

Just as an additional piece of information we also present the Pearson correlation coefficients. They characterize the shape of the relationship (whether it is almost linear or not). The results are presented in Tables 10 and 11. They are obtained for $n=4$. From these tables we can learn, for example, that the Spearman rank correlation between quantities presented in Fig 1. plot A (i.e. between quintiles of order 0.1 and the mean values of SI in its classes $CI_i$) equals 0.729 while the analogous rank correlation computed for index ATI equals 1 (i.e. for data illustrate in Fig. 2 Plot A). Moreover, we can also see that in case of *each of the considered* four types of estimation errors, as well as in the case of each its statistical characteristics, the highest (among all considered indices) rank correlation is gained by the ATI. It proves that this new index is very competitive as an indicator of the PV estimation quality.

Very similar results are obtained for any greater number of alternatives. We perform our studies for $n=4,...,9$. In each case the index ATI turns out to be the best indicator of the PV estimation quality (irrespective of the considered type of error). It is also worth mentioning here that for $n>5$ all Spearman rank correlation coefficients computed for the Index ATI equal 1 (!). Another interesting finding is that in the case of all considered inconsistency indices the greater the number of alternatives, the greater value of the the rank correlation coefficient (with any of considered statistical characteristic). We have already noticed this phenomenon in Tables 2 and 3 presenting results received under the NEE-SF.

**Table 10**. Average Spearman rank correlation coefficients between the mean values of an inconsistency index computed in its classes $IC_i$ ($i=1,...,15$) and the indicated statistical characteristics of PV estimation errors corresponding to these classes. The presented results are computed in the case of $n=4$ on the basis of 240 000 PCMs randomly generated according the MSOBE-SF**.**

|  |  | SI | GI | KI | ATI | SI | GI | KI | ATI |
|---|---|---|---|---|---|---|---|---|---|
|  |  | Correlations with the errors AE(REV) | | | | Correlations with the errors RE(REV) | | | |
| Quantiles | Mean | 0.536 | 0.525 | 0.921 | 1 | 0.935 | 0.957 | 0.954 | 0.989 |
| | $p=0.1$ | 0.729 | 0.754 | 0.961 | 1 | 0.696 | 0.704 | 0.950 | 1 |
| | Median | 0.729 | 0.743 | 0.964 | 1 | 0.989 | 0.996 | 0.982 | 1 |
| | $p=0.9$ | 0.421 | 0.411 | 0.871 | 0.868 | 0.996 | 0.996 | 0.971 | 0.996 |
|  |  | Correlations with the errors AE(GM) | | | | Correlations with the errors RE(GM) | | | |
| Quantiles | Mean | 0.754 | 0.771 | 0.982 | 1 | 0.928 | 0.946 | 0.946 | 0.982 |
| | $p=0.1$ | 0.796 | 0.821 | 0.964 | 1 | 0.964 | 0.982 | 0.975 | 1 |
| | Median | 0.793 | 0.854 | 0.979 | 1 | 0.667 | 0.671 | 0.964 | 1 |
| | $p=0.9$ | 0.496 | 0.500 | 0.939 | 0.996 | 0.953 | 0.961 | 0.961 | 0.993 |

**Table 11**. Average Pearson correlation coefficients between the mean values of the inconsistency index computed in its classes $IC_i$ ($i=1,...,15$) and the indicated statistical characteristics of PV estimation errors corresponding to these classes. The presented results are computed in case of $n=4$ on the basis of 240 000 PCMs randomly generated according the MSOBE-SF**.**

|  |  | SI | GI | KI | ATI | SI | GI | KI | ATI |
|---|---|---|---|---|---|---|---|---|---|
|  |  | Correlations with the errors AE(REV) | | | | Correlations with the errors RE(REV) | | | |
| Quantiles | Mean | 0.510 | 0.557 | 0.938 | 0.943 | 0.947 | 0.925 | 0.612 | 0.723 |
| | $p=0.1$ | 0.615 | 0.656 | 0.947 | 0.978 | 0.819 | 0.839 | 0.935 | 0.979 |
| | Median | 0.587 | 0.630 | 0.953 | 0.953 | 0.927 | 0.906 | 0.738 | 0.820 |
| | $p=0.9$ | 0.360 | 0.413 | 0.896 | 0.890 | 0.903 | 0.874 | 0.550 | 0.663 |
|  |  | Correlations with the errors AE(GM) | | | | Correlations with the errors RE(GM) | | | |
| Quantiles | Mean | 0.803 | 0.827 | 0.976 | 0.981 | 0.975 | 0.968 | 0.737 | 0.832 |
| | $p=0.1$ | 0.636 | 0.685 | 0.953 | 0.974 | 0.734 | 0.775 | 0.961 | 0.986 |
| | Median | 0.733 | 0.769 | 0.978 | 0.981 | 0.903 | 0.904 | 0.930 | 0.944 |
| | $p=0.9$ | 0.851 | 0.850 | 0.909 | 0.953 | 0.922 | 0.897 | 0.589 | 0.703 |

## 5. Final remarks and conclusions

The index ATI turns out to be a very good indicator of the trustworthiness of the PCM as a source of information about the PV. As we have seen when analyzing the Tables 10 and 11 the index ATI performs really well - certainly it performs best among all analyzed indices here. This fact can be also observed when we study the performance of the indices on the basis of various subsets of PCMs, e.g. generated with specific distribution of the small error ε (separately for the gamma, log-normal, truncated normal, and uniform p.d.). It is also confirmed under different simulation frameworks (e.g. we have seen it in Tables 2 and 3). Moreover, the dominance of the index ATI can be observed regardless of the considered number of alternatives, the loss function (AE or RE) and the prioritization method (REV or GM). All these results are essentially the same for a different number $N_C$ of the index classes $IC_i$, $i=1,...,N_C$. The results presented in the last section are received for $N_C=15$, but we have also analyzed the rank coefficients for $N_C=10, 25$, and $50$.

The trustworthy and monotonic relationship between the values of ATI and the quantiles is of special interest to us. It is because these quantiles are likely to be used in order to accept or reject PCM as a good source of information - they tell us about the chances that the estimation error is greater than (or less than) the quantiles' values. It is a fundamental knowledge required in the process of PCM acceptance. That is because one may need to take into account both chances: the chances of small error (i.e. the possibility that given PCM is good, "consistent" matrix) and the chances of big estimates errors (i.e. that we deal with an unacceptable PCM). And a priori we are not at the position to claim which type of estimation error magnitude (small or big) are more important for the decisions about matrix acceptance - it certainly depends on the problem itself and the decision maker's attitude to these errors. Consequently, both quintiles (of order 0.1 and 0.9) of the error magnitude may be important in various specific situations. Which quintile should be used depends on which type of error is regarded as more important. Or, in other words, which is more costly: to reject a "good" matrix or to accept a "bad" one - situation that is well known from classical theory of hypothesis testing. To enable such analysis we provide tables (in the Appendix) which contain statistical characteristics of the estimation errors (namely RE(REV) and RE(GM)) in relation to the ATI values. They are organized in a similar way as Table 9. The results presented in these tables are based on data gathered under the MSOBE-SF performed for $n=4,5,6,7$ in a case where the judgment errors (both type: small and big) were generated in a way described in Section 4. Obviously, as based on simulation data, all the presented statistical characteristics are only the estimates of the true error distribution parameters. Nonetheless they can be helpful while making decisions concerning the acceptance of the PCM at hand. They can also be possibly used for a comparison with results received under different simulation frameworks (e.g. involving different judgment error' distribution).

Finally we state some remarks about the remaining here-considered indices. Although the index KI for $n=4$ performs really well, see Table 10, it is the worst one in indicating PV estimation quality when the number of alternatives $n$ is greater than 5. It may be a result of vanishing influence of the single big error implemented in our simulations. The indices GI and SI perform similarly to each other for $n=4,...,9$. On the other hand, as it was noticed before, all here-considered rank correlations increase when the number *n* increases, and finally the differences between all considered inconsistency indices practically vanish for *n* > 7.

All simulation frameworks described in this paper as well as all computations were coded and run in *Wolfram Mathematica* 9.0

 

**Appendix**. **Tables with the statistical characteristics of PV estimation error distributions in relation to ATI values.**

The tables provided in this appendix contain $p$-quantiles ($p$= 0.1 , 0.5, 0.9) and the arithmetic mean of the random relative errors RE related to different classes $IC_i$, $i$=1,...,15 of the ATI values. In each case the results are based on 240 000 random reciprocal PCMs generated under the MSOBE-SF. The perturbation factors $\varepsilon_{ij}$ was generated -in equal proportions- according to gamma, log-normal, truncated normal, and uniform distributions. The one big error has uniform distribution on the interval [2,4]. In 1/4 of the simulation runs the Step 2 of MSOBE-SF is omitted (meaning that 25% records in the simulation database is related to cases with no "big errors").

Tables A1 - A4 show the results for the REV prioritization method, Tables A5-A8 contain results related to the GM.

**Table A1.** Statistical characteristics of random errors RE(REV) related to different classes of ATI. Results for $n$=4.

| $i$ | classes $IC_i$ of SI | Mean ATI in $IC_i$ | $p$-quantiles in $SR_i$(REV): $p$=0.1 | $p$=0.5 | $p$=0.9 | Mean AE in $SR_i$(REV) |
|---|---|---|---|---|---|---|
| 1 | 0.000 ÷ 0.173 | 0.1111 | 0.0322 | 0.0714 | 0.1765 | 0.1248 |
| 2 | 0.173 ÷ 0.207 | 0.1867 | 0.0376 | 0.0874 | 0.1995 | 0.1480 |
| 3 | 0.207 ÷ 0.240 | 0.2230 | 0.0489 | 0.1135 | 0.2259 | 0.1369 |
| 4 | 0.240 ÷ 0.274 | 0.2614 | 0.0492 | 0.1189 | 0.2762 | 0.2106 |
| 5 | 0.274 ÷ 0.308 | 0.2895 | 0.0524 | 0.1323 | 0.2647 | 0.1883 |
| 6 | 0.308 ÷ 0.341 | 0.3251 | 0.0733 | 0.1576 | 0.3046 | 0.2186 |
| 7 | 0.341 ÷ 0.375 | 0.3586 | 0.0813 | 0.1692 | 0.3545 | 0.2566 |
| 8 | 0.375÷0.409 | 0.3918 | 0.0842 | 0.1776 | 0.3918 | 0.2755 |
| 9 | 0.409÷0.443 | 0.4260 | 0.0885 | 0.1861 | 0.4314 | 0.3738 |
| 10 | 0.443÷0.476 | 0.4593 | 0.0921 | 0.1945 | 0.4438 | 0.3326 |
| 11 | 0.476÷0.510 | 0.4925 | 0.0963 | 0.1986 | 0.5075 | 0.4032 |
| 12 | 0.510÷0.544 | 0.5264 | 0.0989 | 0.2087 | 0.6167 | 0.5741 |
| 13 | 0.544÷0.577 | 0.5597 | 0.1065 | 0.2285 | 0.9665 | 0.7165 |
| 14 | 0.577÷0.611 | 0.5930 | 0.1157 | 0.2590 | 1.4399 | 0.8798 |
| 15 | 0.611÷∞ | 0.6738 | 0.1496 | 0.5685 | 4.9532 | 2.6707 |

**Table A2.** Statistical characteristics of random errors RE(REV) related to different classes of ATI. Results for $n=5$.

| $i$ | classes $IC_i$ of SI | Mean ATI in $IC_i$ | $p$-quantiles in $SR_i$(REV): | | | Mean AE in $SR_i$(REV) |
|---|---|---|---|---|---|---|
| | | | $p=0.1$ | $p=0.5$ | $p=0.9$ | |
| 1 | 0.000 ÷ 0.188 | 0.1403 | 0.0393 | 0.0781 | 0.1537 | 0.1131 |
| 2 | 0.188 ÷ 0.213 | 0.2013 | 0.0455 | 0.0927 | 0.1830 | 0.1419 |
| 3 | 0.213 ÷ 0.237 | 0.2252 | 0.0502 | 0.1026 | 0.2002 | 0.1623 |
| 4 | 0.237 ÷ 0.262 | 0.2498 | 0.0577 | 0.1152 | 0.2234 | 0.1607 |
| 5 | 0.262 ÷ 0.286 | 0.2743 | 0.0622 | 0.1261 | 0.2526 | 0.1967 |
| 6 | 0.286 ÷ 0.311 | 0.2988 | 0.0682 | 0.1334 | 0.2764 | 0.2078 |
| 7 | 0.311 ÷ 0.335 | 0.3231 | 0.0732 | 0.1411 | 0.2958 | 0.2183 |
| 8 | 0.335÷0.360 | 0.3476 | 0.0779 | 0.1499 | 0.3476 | 0.2610 |
| 9 | 0.360÷0.384 | 0.3720 | 0.0816 | 0.1595 | 0.4107 | 0.2868 |
| 10 | 0.384÷0.409 | 0.3959 | 0.0869 | 0.1693 | 0.4974 | 0.3922 |
| 11 | 0.409÷0.433 | 0.4206 | 0.0907 | 0.1835 | 0.6375 | 0.4050 |
| 12 | 0.433÷0.458 | 0.4448 | 0.0975 | 0.2081 | 0.9019 | 0.5648 |
| 13 | 0.458÷0.482 | 0.4692 | 0.1061 | 0.2490 | 1.3716 | 0.7300 |
| 14 | 0.482÷0.507 | 0.4939 | 0.1185 | 0.3192 | 1.9143 | 1.0793 |
| 15 | 0.507 ÷ ∞ | 0.5633 | 0.1694 | 0.8982 | 5.6261 | 3.6842 |

**Table A3.** Statistical characteristics of random errors RE(REV) related to different classes of ATI. Results for $n=6$.

| $i$ | classes $IC_i$ of SI | Mean ATI in $IC_i$ | $p$-quantiles in $SR_i$(REV): | | | Mean AE in $SR_i$(REV) |
|---|---|---|---|---|---|---|
| | | | $p=0.1$ | $p=0.5$ | $p=0.9$ | |
| 1 | 0.000 ÷ 0.194 | 0.1570 | 0.0428 | 0.0789 | 0.1440 | 0.1007 |
| 2 | 0.194 ÷ 0.214 | 0.2043 | 0.0490 | 0.0918 | 0.1656 | 0.1247 |
| 3 | 0.214 ÷ 0.234 | 0.2244 | 0.0543 | 0.0987 | 0.1782 | 0.1396 |
| 4 | 0.234 ÷ 0.254 | 0.2444 | 0.0583 | 0.11052 | 0.2004 | 0.1744 |
| 5 | 0.254 ÷ 0.274 | 0.2643 | 0.0620 | 0.1126 | 0.2256 | 0.1980 |
| 6 | 0.274 ÷ 0.294 | 0.2842 | 0.0663 | 0.1194 | 0.2615 | 0.2132 |
| 7 | 0.294 ÷ 0.314 | 0.3040 | 0.0706 | 0.1284 | 0.3042 | 0.2224 |
| 8 | 0.314÷0.334 | 0.3239 | 0.0751 | 0.1392 | 0.3654 | 0.2702 |
| 9 | 0.334÷0.354 | 0.3438 | 0.0799 | 0.1507 | 0.4493 | 0.3493 |
| 10 | 0.354÷0.374 | 0.3638 | 0.0866 | 0.1690 | 0.5996 | 0.4314 |
| 11 | 0.374÷0.394 | 0.3837 | 0.0921 | 0.1922 | 0.7899 | 0.5753 |
| 12 | 0.394÷0.414 | 0.4035 | 0.0992 | 0.2259 | 1.0904 | 0.7326 |
| 13 | 0.414÷0.434 | 0.4235 | 0.1099 | 0.3090 | 1.7322 | 1.2208 |
| 14 | 0.434÷0.454 | 0.4434 | 0.1210 | 0.4270 | 2.2000 | 1.4294 |
| 15 | 0.454 ÷ ∞ | 0.5009 | 0.1801 | 0.9724 | 6.4509 | 3.2361 |

**Table A4.** Statistical characteristics of random errors RE(REV) related to different classes of ATI. Results for $n=7$.

| | classes $IC_i$ | Mean ATI | $p$-quantiles in $SR_i$(REV): | Mean AE in |
|---|---|---|---|---|

| $i$ | classes $IC_i$ of SI | Mean ATI in $IC_i$ | $p$-quantiles in $SR_i$(GM): | | | Mean AE in $SR_i$(REV) |
|---|---|---|---|---|---|---|
| | | | $p=0.1$ | $p=0.5$ | $p=0.9$ | |
| 1 | 0.000 ÷ 0.194 | 0.162 | 0.0453 | 0.0764 | 0.1350 | 0.1033 |
| 2 | 0.194 ÷ 0.212 | 0.204 | 0.0499 | 0.0860 | 0.1548 | 0.1288 |
| 3 | 0.212 ÷ 0.229 | 0.221 | 0.0535 | 0.0922 | 0.1719 | 0.1442 |
| 4 | 0.229 ÷ 0.247 | 0.238 | 0.0574 | 0.0994 | 0.1934 | 0.1562 |
| 5 | 0.247 ÷ 0.265 | 0.256 | 0.0613 | 0.1056 | 0.2156 | 0.1842 |
| 6 | 0.265 ÷ 0.282 | 0.273 | 0.0652 | 0.1139 | 0.2463 | 0.2064 |
| 7 | 0.282 ÷ 0.300 | 0.291 | 0.0689 | 0.1223 | 0.3074 | 0.2404 |
| 8 | 0.300 ÷ 0.317 | 0.308 | 0.0731 | 0.1327 | 0.3771 | 0.2836 |
| 9 | 0.317 ÷ 0.335 | 0.326 | 0.0786 | 0.1489 | 0.4867 | 0.3341 |
| 10 | 0.335 ÷ 0.353 | 0.343 | 0.0863 | 0.1734 | 0.6389 | 0.4274 |
| 11 | 0.353 ÷ 0.370 | 0.361 | 0.0918 | 0.2090 | 0.8335 | 0.5286 |
| 12 | 0.370 ÷ 0.388 | 0.378 | 0.1025 | 0.2651 | 1.1082 | 0.6795 |
| 13 | 0.388 ÷ 0.405 | 0.396 | 0.1144 | 0.3563 | 1.4908 | 0.8430 |
| 14 | 0.405 ÷ 0.423 | 0.413 | 0.1280 | 0.4644 | 2.1353 | 1.2980 |
| 15 | 0.423 ÷ ∞ | 0.466 | 0.1944 | 1.0171 | 7.5328 | 3.4828 |

**Table A5.** Statistical characteristics of random errors RE(GM) related to different classes of ATI. Results for $n=4$.

| $i$ | classes $IC_i$ of SI | Mean ATI in $IC_i$ | $p$-quantiles in $SR_i$(GM): | | | Mean AE in $SR_i$(GM) |
|---|---|---|---|---|---|---|
| | | | $p=0.1$ | $p=0.5$ | $p=0.9$ | |
| 1 | 0.000 ÷ 0.173 | 0.1111 | 0.0324 | 0.0700 | 0.1780 | 0.1242 |
| 2 | 0.173 ÷ 0.207 | 0.1867 | 0.0371 | 0.0836 | 0.1939 | 0.1460 |
| 3 | 0.207 ÷ 0.240 | 0.2230 | 0.0469 | 0.1144 | 0.2182 | 0.1342 |
| 4 | 0.240 ÷ 0.274 | 0.2614 | 0.0488 | 0.1195 | 0.2688 | 0.2029 |
| 5 | 0.274 ÷ 0.308 | 0.2895 | 0.0516 | 0.1281 | 0.2576 | 0.1834 |
| 6 | 0.308 ÷ 0.341 | 0.3251 | 0.0805 | 0.1548 | 0.2879 | 0.2082 |
| 7 | 0.341 ÷ 0.375 | 0.3586 | 0.0883 | 0.1710 | 0.3146 | 0.2433 |
| 8 | 0.375 ÷ 0.409 | 0.3918 | 0.0924 | 0.1844 | 0.3347 | 0.2571 |
| 9 | 0.409 ÷ 0.443 | 0.4260 | 0.1005 | 0.1941 | 0.3551 | 0.3329 |
| 10 | 0.443 ÷ 0.476 | 0.4593 | 0.1101 | 0.1998 | 0.3520 | 0.2964 |
| 11 | 0.476 ÷ 0.510 | 0.4925 | 0.1159 | 0.2017 | 0.3975 | 0.3262 |
| 12 | 0.510 ÷ 0.544 | 0.5264 | 0.1229 | 0.2042 | 0.4681 | 0.4317 |
| 13 | 0.544 ÷ 0.577 | 0.5597 | 0.1261 | 0.2134 | 0.6839 | 0.5168 |
| 14 | 0.577 ÷ 0.611 | 0.5930 | 0.1330 | 0.2215 | 0.9398 | 0.6015 |
| 15 | 0.611 ÷ ∞ | 0.6738 | 0.1527 | 0.3499 | 2.5202 | 1.1846 |

**Table A6.** Statistical characteristics of random errors RE(GM) related to different classes of ATI. Results for $n=5$.

| $i$ | classes $IC_i$ of SI | Mean ATI in $IC_i$ | $p$-quantiles in $SR_i$(GM): | | | Mean AE in $SR_i$(GM) |
|---|---|---|---|---|---|---|
| | | | $p=0.1$ | $p=0.5$ | $p=0.9$ | |
| 1 | 0.000 ÷ 0.188 | 0.1403 | 0.0384 | 0.0758 | 0.1496 | 0.1102 |
| 2 | 0.188 ÷ 0.213 | 0.2013 | 0.0433 | 0.0898 | 0.1762 | 0.1356 |
| 3 | 0.213 ÷ 0.237 | 0.2252 | 0.0492 | 0.0995 | 0.1874 | 0.1515 |
| 4 | 0.237 ÷ 0.262 | 0.2498 | 0.0559 | 0.1108 | 0.2057 | 0.1475 |
| 5 | 0.262 ÷ 0.286 | 0.2743 | 0.0618 | 0.1229 | 0.2194 | 0.1772 |
| 6 | 0.286 ÷ 0.311 | 0.2988 | 0.0689 | 0.1296 | 0.2328 | 0.1814 |
| 7 | 0.311 ÷ 0.335 | 0.3231 | 0.0745 | 0.1373 | 0.2457 | 0.1932 |
| 8 | 0.335 ÷ 0.360 | 0.3476 | 0.0803 | 0.1449 | 0.2786 | 0.2263 |
| 9 | 0.360 ÷ 0.384 | 0.3720 | 0.0867 | 0.1519 | 0.3171 | 0.2397 |

| $i$ | classes $IC_i$ of SI | Mean ATI in $IC_i$ | $p$=0.1 | $p$=0.5 | $p$=0.9 | Mean AE in $SR_i(GM)$ |
|---|---|---|---|---|---|---|
| 10 | 0.384÷0.409 | 0.3959 | 0.0927 | 0.1559 | 0.3827 | 0.3062 |
| 11 | 0.409÷0.433 | 0.4206 | 0.0964 | 0.1632 | 0.4709 | 0.3148 |
| 12 | 0.433÷0.458 | 0.4448 | 0.1005 | 0.1745 | 0.6285 | 0.4056 |
| 13 | 0.458÷0.482 | 0.4692 | 0.1055 | 0.1936 | 0.9010 | 0.5006 |
| 14 | 0.482÷0.507 | 0.4939 | 0.1133 | 0.2232 | 1.2012 | 0.6908 |
| 15 | 0.507÷∞ | 0.5633 | 0.1406 | 0.5300 | 2.8428 | 1.8051 |

**Table A7.** Statistical characteristics of random errors RE(GM) related to different classes of ATI. Results for $n$=6.

| $i$ | classes $IC_i$ of SI | Mean ATI in $IC_i$ | $p$-quantiles in $SR_i(GM)$: $p$=0.1 | $p$=0.5 | $p$=0.9 | Mean AE in $SR_i(GM)$ |
|---|---|---|---|---|---|---|
| 1 | 0.000 ÷ 0.194 | 0.1570 | 0.0413 | 0.0761 | 0.1362 | 0.0954 |
| 2 | 0.194 ÷ 0.214 | 0.2043 | 0.0461 | 0.0871 | 0.1518 | 0.1148 |
| 3 | 0.214 ÷ 0.234 | 0.2244 | 0.0519 | 0.0943 | 0.1608 | 0.1278 |
| 4 | 0.234 ÷ 0.254 | 0.2444 | 0.0556 | 0.0993 | 0.1745 | 0.1564 |
| 5 | 0.254 ÷ 0.274 | 0.2643 | 0.0600 | 0.1054 | 0.1896 | 0.1695 |
| 6 | 0.274 ÷ 0.294 | 0.2842 | 0.0644 | 0.1116 | 0.2136 | 0.1831 |
| 7 | 0.294 ÷ 0.314 | 0.3040 | 0.0684 | 0.1181 | 0.2426 | 0.1858 |
| 8 | 0.314÷0.334 | 0.3239 | 0.0730 | 0.1247 | 0.2826 | 0.2184 |
| 9 | 0.334÷0.354 | 0.3438 | 0.0771 | 0.1316 | 0.3396 | 0.2688 |
| 10 | 0.354÷0.374 | 0.3638 | 0.0810 | 0.1404 | 0.4366 | 0.3161 |
| 11 | 0.374÷0.394 | 0.3837 | 0.0862 | 0.1525 | 0.5567 | 0.3967 |
| 12 | 0.394÷0.414 | 0.4035 | 0.0906 | 0.1702 | 0.7314 | 0.4822 |
| 13 | 0.414÷ 0.434 | 0.4235 | 0.0975 | 0.2137 | 1.0455 | 0.7569 |
| 14 | 0.434÷ 0.454 | 0.4434 | 0.1062 | 0.2791 | 1.3872 | 0.8573 |
| 15 | 0.454 ÷ ∞ | 0.5009 | 0.1372 | 0.5867 | 3.1325 | 1.6558 |

**Table A8.** Statistical characteristics of random errors RE(GM) related to different classes of ATI. Results for $n$=7.

| $i$ | classes $IC_i$ of SI | Mean ATI in $IC_i$ | $p$-quantiles in $SR_i(GM)$: $p$=0.1 | $p$=0.5 | $p$=0.9 | Mean AE in $SR_i(GM)$ |
|---|---|---|---|---|---|---|
| 1 | 0.000 ÷ 0.194 | 0.162 | 0.0424 | 0.0723 | 0.1225 | 0.945 |
| 2 | 0.194 ÷ 0.212 | 0.204 | 0.0460 | 0.0796 | 0.1381 | 0.1150 |
| 3 | 0.212 ÷ 0.229 | 0.221 | 0.0494 | 0.0850 | 0.1475 | 0.1277 |
| 4 | 0.229 ÷ 0.247 | 0.238 | 0.0542 | 0.0907 | 0.1595 | 0.1360 |
| 5 | 0.247 ÷ 0.265 | 0.256 | 0.0572 | 0.0956 | 0.1734 | 0.1537 |
| 6 | 0.265 ÷ 0.282 | 0.273 | 0.0611 | 0.1011 | 0.1941 | 0.1702 |
| 7 | 0.282 ÷ 0.300 | 0.291 | 0.0640 | 0.1072 | 0.2345 | 0.1916 |
| 8 | 0.300÷ 0.317 | 0.308 | 0.0680 | 0.1136 | 0.2793 | 0.2220 |
| 9 | 0.317÷0.335 | 0.326 | 0.0715 | 0.1226 | 0.3600 | 0.2526 |
| 10 | 0.335÷0.353 | 0.343 | 0.0767 | 0.1351 | 0.4512 | 0.3105 |
| 11 | 0.353÷0.370 | 0.361 | 0.0808 | 0.1556 | 0.5578 | 0.3722 |
| 12 | 0.370÷0.388 | 0.378 | 0.0883 | 0.1884 | 0.7042 | 0.4576 |
| 13 | 0.388÷0.405 | 0.396 | 0.0976 | 0.2367 | 0.9399 | 0.5426 |
| 14 | 0.405÷ 0.423 | 0.413 | 0.1041 | 0.2977 | 1.3014 | 0.8001 |
| 15 | 0.423 ÷ ∞ | 0.466 | 0.1422 | 0.6174 | 3.8187 | 1.8220 |